\documentclass[runningheads]{llncs}
\usepackage{graphicx}
\usepackage{booktabs}
\usepackage{rotating}
\usepackage{multirow}
\usepackage{multicol}
\usepackage{xcolor}
\usepackage{pifont}
\usepackage{amsmath}
\usepackage{amssymb}
\usepackage{hyperref}
	
\usepackage{color, colortbl}
\definecolor{Gray}{gray}{0.9}
\newcommand{\rcl}{\rowcolor{Gray}}
\newcommand{\ccl}{\cellcolor{Gray}}

\newcommand{\yes}{\textcolor{green}{\ding{51}}}
\newcommand{\no}{\textcolor{red}{\ding{55}}}

\begin{document}

\title{Representation learning in multiplex graphs: Where and how to fuse information?}
\titlerunning{Fusion in multiplex graph representation learning}

\author{Piotr Bielak\inst{1}\orcidID{0000-0002-1487-2569} \and
Tomasz Kajdanowicz\inst{1}\orcidID{0000-0002-8417-1012}}


\institute{
Department of Artificial Intelligence,\\
Wroclaw University of Science and Technology,\\
Wrocław, Poland\\
\email{piotr.bielak@pwr.edu.pl}
}

\maketitle

\begin{abstract}
In recent years, unsupervised and self-supervised graph representation learning has gained popularity in the research community. However, most proposed methods are focused on homogeneous networks, whereas real-world graphs often contain multiple node and edge types. Multiplex graphs, a special type of heterogeneous graphs, possess richer information, provide better modeling capabilities and integrate more detailed data from potentially different sources. The diverse edge types in multiplex graphs provide more context and insights into the underlying processes of representation learning. In this paper, we tackle the problem of learning representations for nodes in multiplex networks in an unsupervised or self-supervised manner. To that end, we explore diverse information fusion schemes performed at different levels of the graph processing pipeline. The detailed analysis and experimental evaluation of various scenarios inspired us to propose improvements in how to construct GNN architectures that deal with multiplex graphs.

\end{abstract}

\section{Introduction}
Real-world data often exhibits a complex and heterogeneous nature, leading to challenges and opportunities in graph representation learning. From social networks, where multiple relationship types exist among users, to biological networks, where various interaction types occur among molecules, a multiplex view is often more appropriate to capture the entirety of the network structure. Multiplex networks, characterized by the presence of multiple edge types (graph layers), span across a common set of nodes, provide richer information, better modeling capabilities, and facilitate the integration of more detailed data from diverse sources. As such, a multiplex network is not simply a combination of homogeneous graphs but a fundamentally more complex system encapsulating a range of various interaction types. Despite the rich information inherent to multiplex networks, the research focus in graph representation learning has been overwhelmingly concentrated on methods for homogeneous graphs. This has left a substantial gap in understanding how to leverage the unique features of multiplex networks effectively.

To this end, we examine the problem of learning representations for nodes in multiplex networks from the perspective of unsupervised and self-supervised learning. Unlike supervised learning, these approaches do not require labeled data and, hence, offer a flexible and scalable method for learning node representations. In particular, we investigate various information fusion schemes that can be performed at different stages of the representation learning pipeline to harness the richness of multiplex networks. These schemes are designed to integrate the diverse edge types in multiplex graphs, providing more context and insights into the underlying processes.

This paper details an extensive analysis and experimental evaluation of various information fusion scenarios. We hope our work advances our understanding of multiplex network representation learning and paves the way for developing more robust, efficient, and versatile multiplex network methods.

Our contributions can be summarized as follows:
\begin{itemize}
    \item[1.] We propose an information fusion taxonomy for multiplex networks.
    \item[2.] We provide an extensive experimental evaluation of existing representation learning approaches in multiplex networks. We categorize them according to our proposed taxonomy. We evaluate the node embeddings in three tasks: node classification, node clustering and similarity search.
    \item[3.] We identify research gaps in each fusion strategy category and provide preliminary implementations of those methods. We evaluate their performance and compare them to existing approaches. We provide a limitations analysis of existing methods and identify future directions in representation learning for multiplex networks.
\end{itemize}

\section{Related work}
\paragraph{\textbf{Graph representation learning.}} It has emerged as a vital area of research with applications spanning various domains, such as social networks, bioinformatics, or recommender systems. The essence lies in learning continuous, low-dimensional representations of nodes that capture their structural and attribute information within the graph. Plenty of methods have been developed, each focused on a different aspect of network embeddings, such as structure, attributes, learning paradigm or scalability. Shallow methods, such as DeepWalk \cite{Perozzi:2014:DOL:2623330.2623732}, LINE \cite{tang2015line} use a simple notion of graph coding through random walks or objectives that optimize first and second-order node similarity. More complex graph neural networks, such as GCN\cite{kipf2017semi} or GAT \cite{velickovic2018graph}, implement the message-passing algorithm over graph edges combined with various message aggregation schemes.

\paragraph{\textbf{Un- and self-supervised graph representation learning.}} Inspired by the success of contrastive methods in other domains, the procedures were adapted to graphs. Early approaches were based on the autoencoder architecture (GAE \cite{kipf2016variational}). Another method, DGI \cite{velickovic2018deep}, employed a GNN to learn node embeddings and maximized the mutual information between node embeddings and the graph embedding (readout function) by discriminating nodes in the original graph from nodes in a corrupted graph. GRACE \cite{Zhu:2020vf} and GraphCL \cite{You2020GraphCL} utilized contrastive learning on graphs. All the previous methods rely on negative sampling which yields a high complexity. Negative-sample-free methods, such as BGRL \cite{thakoor2021bootstrapped} or GBT \cite{BIELAK2022109631} use graph augmentations combined with either an asymmetric pipeline architecture or a decorrelation-based loss to prevent representation collapse and learn node embeddings in a self-supervised manner.

\paragraph{\textbf{Multiplex graph representation learning.}} Despite the advances in representation learning for homogeneous graphs, research on multiplex graphs is relatively limited. MHGCN \cite{mhgcn} utilizes a weighted sum of adjacency matrices of the multiplex graph layers, where the combination weights are learnable. The resulting graph is passed into a standard GCN backbone while the whole model is trained using a link prediction objective with negative sampling. DMGI \cite{dmgi}, extends the idea of DGI to multiplex graphs. It applies a GCN backbone with the DGI objective for each graph layer. Node embeddings are fused using a trainable lookup embedding that optimized via a loss function that minimizes the distance layer-wise node embeddings of the original graph and maximizes the distance to embeddings from the corrupted graph. HDGI \cite{hdgi} also builds upon the DGI method. It applies a GCN backbone at each graph layer and fuses the resulting vectors into a single node embedding using a semantic attention mechanism. Such scheme is applied to the original graph and its corrupted version. The two node embedding sets are processed by DGI's original loss function. Contrary to previous approaches, S$^2$MGRL \cite{s2mgrl} extends the idea of GBT \cite{BIELAK2022109631} to multiplex networks. For each graph layer, the model first applies an MLP followed by a GCN block. The loss function applies the Barlow Twins objective \cite{zbontar2021barlow} between (a) the outputs of the MLPs and GCNs, and (b) all pairs of GCN outputs to ensure correlation between embeddings from all graph layers. This training step (and all previously introduced multiplex graph representation learning approaches) is carried out in an un- or self-supervised manner. However, S$^2$MGRL trains an attention mechanism on the frozen embedding from all layers to obtain the final node embeddings, utilizing node labels on the output.

\section{Information fusion in multiplex graphs}
\begin{figure}[ht]
    \centering
    \includegraphics[width=\textwidth]{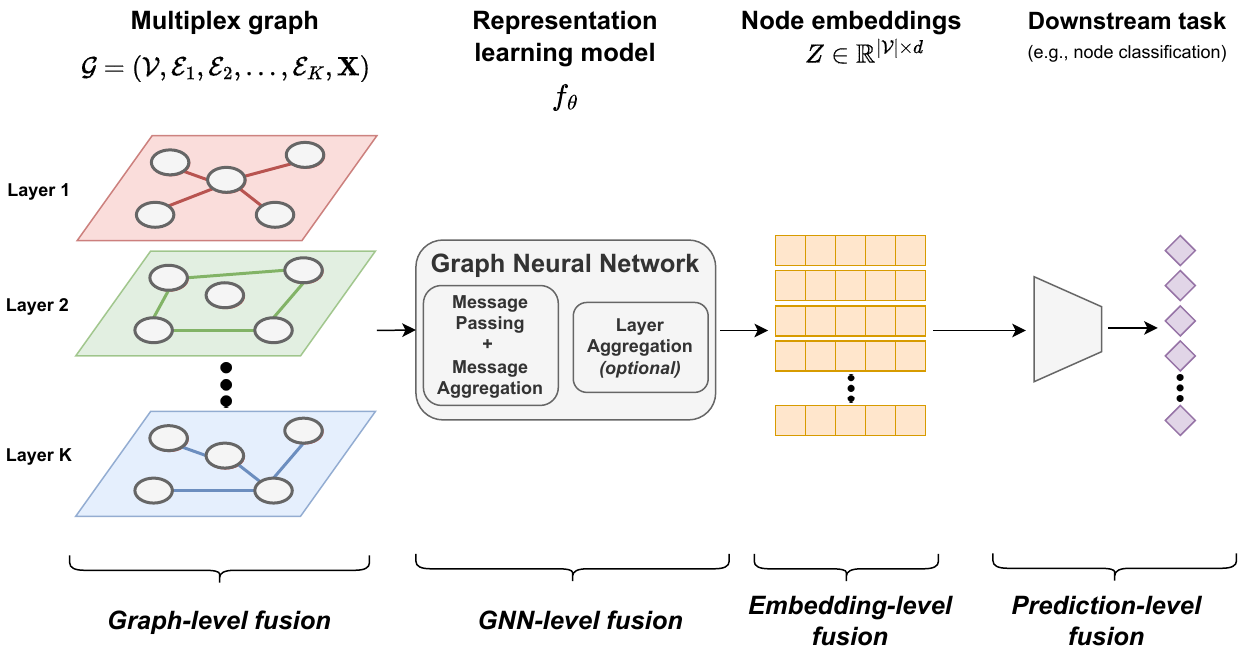}
    \caption{Multiplex graph processing pipeline and information fusion taxonomy.}
    \label{fig:multiplex-fusion-pipeline}
\end{figure}

Let us start with basic definitions and the introduction of the multiplex graph representation learning pipeline (see: Figure \ref{fig:multiplex-fusion-pipeline}). Next, we will introduce a taxonomy of information fusion methods in multiplex graphs, followed by a detailed discussion of existing methods from each category and our proposed extensions.

\paragraph{\textbf{Multiplex graph.}} We define a multiplex graph $\mathcal{G} = (\mathcal{V}, \mathcal{E}_1, \mathcal{E}_2, \ldots, \mathcal{E}_K, \mathbf{X})$, where $\mathcal{V}$ denotes a set of nodes, $\mathbf{X} \in \mathbb{R}^{|\mathcal{V}| \times d_\text{in}}$ is a matrix whose rows contain $d_\text{in}$-dimensional node's features. The $K$ layers of the multiplex graph are defined by the $K$ sets of edges: $\mathcal{E}_1, \mathcal{E}_2, \ldots, \mathcal{E}_K$, where each  set $\mathcal{E}_i \in \mathcal{V} \times \mathcal{V}$ contains edges between pairs of nodes. Note that the nodes and their features are shared across all graph layers.

\paragraph{\textbf{Multiplex graph representation learning.}} When learning embeddings for nodes in multiplex graphs, the input graph is processed by a representation learning model $f_\theta$ (parametrized by $\theta$), which computes embedding vectors $\mathbf{Z} \in \mathbb{R}^{|\mathcal{V}| \times d}$ for all nodes in the graph. This representation learning model is often a graph neural network (GNN) based on the message-passing paradigm. In particular, each graph neural network layer is comprised of two phases -- message generation and message aggregation (see Equation \ref{eq:message-passing}):

\begin{align}
    \label{eq:message-passing}
    \begin{split}
    \mathbf{m}_i = \text{MESSAGE}(\mathbf{h}^{(in)}_i) \\
    \mathbf{h}_u^{(out)} = \text{UPDATE}\left(\mathbf{h}_u^{(in)}, \text{AGG}\left(\{\mathbf{m}_v, v \in \mathcal{N}(u)\}\right)\right)
    \end{split}
\end{align}

where $\text{MESSAGE}(\cdot)$ is a function that converts node $i$'s input features $h^{(in)}_i$ into a message vector $\mathbf{m}_i$ (for the first GNN layer, we use $H^{(in)} = \mathbf{X}$). Incoming messages from node $u$'s neighbors $\mathcal{N}(u)$ are aggregated using a permutation-invariant function $\text{AGG}(\cdot)$. Finally, the $u$'s output representation (embedding) is computed via the $\text{UPDATE}(\cdot)$ function, which takes into account both the node's input features $h^{(in)}_u$ and the aggregated neighbor messages.

GNN layers are modified to accommodate the multiple edge types found in multiplex graphs, or separate GNNs are applied on each multiplex graph layer. In the latter case, which is often used in existing multiplex representation learning methods, the output of those GNNs must be fused into a single representation vector for each node.

The resulting node embedding vectors can be applied in various downstream tasks, such as node classification or link prediction. For instance, a node classifier can be trained in a supervised manner on the frozen node representation vectors and corresponding labels. 

\paragraph{\textbf{Multiplex graph information fusion taxonomy.}} The complex and rich structure of multiplex graphs and their representation learning and processing pipeline enables us to identify several locations for applying information fusion. We propose the following taxonomy of information fusion in multiplex graphs:
\begin{itemize}
    \item \textbf{Graph-level fusion} -- methods that directly modify the graph structure and fuse the information from several multiplex graph layers together,
    \item \textbf{GNN-level fusion} -- methods that either utilize a dedicated GNN architecture that is specifically designed for multiplex graphs or models that utilize standard GNNs in combination with trainable fusion mechanisms (such as attention) to produce fused embedding vectors for nodes,
    \item \textbf{Embedding-level fusion} -- methods that firstly precompute node embeddings in each multiplex graph layer and then fuse the frozen embeddings into a single embedding per node (post-hoc methods),  
    \item \textbf{Prediction-level fusion} -- methods that utilize precomputed node embeddings at each multiplex graph layer and then train downstream task models to obtain and fuse their predictions; note that these methods are not part of the unsupervised or self-supervised representation learning pipeline.
\end{itemize}

Such taxonomy can capture various multiplex graph representation learning methods and categorize them according to where the critical information fusion step is applied. Moreover, one could build another categorization based on the actual fusion mechanism. However, to keep things clearer, we introduce them in detail in Sections \ref{sec:graph-level-fusion-section}-\ref{sec:embedding-level-fusion-section}.

\subsection{Experimental setup and evaluation}
In the following sections, we will introduce the existing methods for each category in our proposed taxonomy, along with an experimental evaluation and comparison of the methods' performance in selected downstream tasks. We implemented all models using the PyTorch-Geometric \cite{Fey/Lenssen/2019} library, and we utilize the DVC package \cite{ruslan_kuprieiev_2021_4733984} to ensure reproducibility. We make our code and data publicly available at \url{https://github.com/graphml-lab-pwr/multiplex-fusion}. Due to page limits, we leave the details about the hyperparameter search ranges and the final chosen ones in the code repository. Let us now introduce the utilized datasets and downstream tasks with the appropriate evaluation protocols.

    

\begin{table}[ht]
    \centering
    \caption{Dataset statistics. We summarize the number of nodes ($|\mathcal{V}|$), the names of particular graph layers along with the number of edges in each layer ({\tiny(\textbf{Layer $i$})} $|\mathcal{E}_i|$), the node feature dimension ($d_\text{in}$) and the number of classes in the node classification task.}
    \label{tab:dataset-statistics}
    
    \resizebox{\textwidth}{!}{
    \begin{tabular}{l|rrr rrr}
        \toprule
        & \textbf{ACM} & \textbf{Amazon} & \textbf{Freebase} & \textbf{IMDB} & \textbf{Cora} & \textbf{CiteSeer} \\
        \midrule

        $|\mathcal{V}|$ & 3,025 & 7,621 & 3,492 & 3,550 & 2,708 & 3,327 \\
        
        {\tiny(\textbf{Layer $i$})} $|\mathcal{E}_i|$ 
        & {\tiny (PAP)} 29,281
        & {\tiny (IBI)} 1,104,257
        & {\tiny (MAM)} 254,702
        & {\tiny (MAM)} 66,428
        & {\tiny (CIT)} 10,556
        & {\tiny (CIT)} 9,104 \\

        & {\tiny (PSP)} 2,210,761
        & {\tiny (IOI)} 14,305
        & {\tiny (MDM)} 8,404
        & {\tiny (MDM)} 13,788
        & {\tiny (KNN)} 54,160
        & {\tiny (KNN)} 66,540 \\

        & 
        & {\tiny (IVI)} 266,237
        & {\tiny (MWM)} 10,706
        &
        &
        & \\

        $d_\text{in}$
        & 1,870
        & 2,000
        & 3,492
        & 2,000
        & 1,433
        & 3,703 \\

        \#classes
        & 3
        & 4
        & 3
        & 3
        & 7
        & 6 \\
        \bottomrule
    \end{tabular}
    }
\end{table}

\paragraph{\textbf{Datasets.}} We selected six real-world datasets frequently utilized in representation learning papers for multiplex graphs \cite{dmgi,hdgi,s2mgrl,mhgcn}. We summarize the statistics of those datasets in Table \ref{tab:dataset-statistics}.
\begin{itemize}
    \item ACM \cite{dataset-acm} is an academic paper dataset with nodes assigned to one of 3 classes: database, wireless communication, and data mining. The multiplex layers are built from 2 meta-paths obtained from this graph: Paper-Author-Paper (PAP) and Paper-Subject-Paper (PSP). Node features are TF-IDF encodings of the papers' keywords. Note that the PSP layer has significantly more edges than the PAP layer.
    \item Amazon \cite{dataset-amazon} is a multiplex network dataset of items bought on Amazon. The graph layers refer to common activities of users, i.e., also-bought (IBI), bought-together (IOI) and also-viewed (IVI). Each item can be categorized into one of 4 classes – Beauty, Automotive, Patio Lawn and Garden, and Baby. Note that there is a disproportion in the size of particular graph layers. Node features are built as bag-of-words encodings of the item descriptions.
    \item Freebase \cite{dataset-freebase} is a movies graph dataset (from the Freebase KG) with nodes divided into three classes -- Action, Comedy and Drama. Graph layers are built upon the following meta-paths: Movie-Actor-Movie (MAM), Movie-Director-Movie (MDM) and Movie-Writer-Movie (MWM). No inherent node features are available, so a one-hot encoding is utilized.
    \item IMDB \cite{dataset-imdb} is another movie graph dataset extracted from the IMDB knowledge graph. Nodes are movies, and each is assigned one of 3 classes – Action, Comedy and Drama. Graph layers are built upon the following meta-paths: Movie-Actor-Movie (MAM) and Movie-Director-Movie (MDM). The node features are TF-IDF encodings of movie metadata, such as title or language.
    \item Cora, CiteSeer \cite{kipf2017semi} are two citation graphs where nodes are research papers divided into seven and six different research areas, respectively. Node features are bag-of-words encodings of paper abstracts. These datasets are inherently homogeneous, and the original edges are put into the CIT layer. We build the multiplex graph by extending the original networks by another graph layer –– a paper similarity layer. We find each node's k-nearest neighbors (KNN) based on the cosine similarity of node features and select the top 10 similar papers (i.e., $k = 10$).
\end{itemize}

\paragraph{\textbf{Downstream tasks.}} The utilized datasets are equipped with node labels, so a node classification is a natural choice for the downstream evaluation of node representations. However, to fully explore the generalization abilities of the learned embedding vectors, we employ two additional tasks -- node clustering and node similarity search \cite{dmgi}. For all tasks, we collect various metrics (provided in the code repository); however, to ensure clarity, we focus only on one metric per task in the paper's result tables. Before executing the downstream evaluation, we freeze the representation backbone model and extract the node embeddings.

\begin{itemize}
    \item \underline{node classification} (\textbf{Clf}) -- we follow the standard evaluation protocol for unsupervised and self-supervised node representation learning methods \cite{velickovic2018deep}. We train a logistic regression classifier (Scikit-learn \cite{scikit-learn}) on the node embeddings and corresponding label pairs from the training set. We provide the Macro-F1 (\textbf{MaF1}) metric on the test set. Additionally, we use the this task for the models' hyperparameter searches. We select the models with the highest Macro-F1 score on the validation set.
    
    \item \underline{node clustering} (\textbf{Clu}) -- we train a K-Means model on the node embeddings and measure the clustering performance using the Normalized Mutual Information (\textbf{NMI}) score on the test set. We train the clustering model ten times with different seeds and average the scores to obtain the final NMI score.
    
    \item \underline{node similarity search} -- we compute the cosine similarity scores of the node embeddings between all pairs of nodes and build a ranking for each node according to the similarity score. Next, we calculate the ratio of the nodes that belong to the same class within top-5 ranked nodes (\textbf{Sim@5}). 
\end{itemize}

\paragraph{\textbf{Table notation.}} In all result tables (Tab. \ref{tab:no-fusion-results}-\ref{tab:method-comparison}), we utilize the same notations and present the mean and std (given in parentheses) of the Macro-F1, NMI and Sim@5 scores as percentages in the node classification, node clustering and similarity search tasks, respectively. Please refer to the \textbf{\textit{Methods.}} paragraphs in corresponding sections for explanations of model names. Shaded rows denote our proposed methods and extensions.

\subsection{Baseline approaches (no fusion)\label{sec:baseline-approches}}

\paragraph{\textbf{Methods.}} We examine two baseline approaches -- \texttt{Layers} and \texttt{Features}. In the first approach, we train the DGI \cite{velickovic2018deep} model on each graph layer separately and evaluate the resulting embeddings. We selected DGI due to its popularity in multiplex representation learning method design and its ability to learn node embeddings in an unsupervised way. The \texttt{Features} approach does not involve any learning but evaluates the initial node features as the node representations. 

\begin{table}[ht]
    \centering

    \caption{Downstream tasks performance of methods \textbf{without information fusion}.} 
    \label{tab:no-fusion-results}

\resizebox{\textwidth}{!}{
\begin{tabular}{l|ccc|ccc|ccc}
        \toprule
        & \multicolumn{3}{c|}{\textbf{ACM}} 
        & \multicolumn{3}{c|}{\textbf{Amazon}}
        & \multicolumn{3}{c}{\textbf{Freebase}} \\
        
        & \textbf{Clf} (MaF1) & \textbf{Clu} (NMI) & \textbf{Sim@5}
        & \textbf{Clf} (MaF1) & \textbf{Clu} (NMI) & \textbf{Sim@5}
        & \textbf{Clf} (MaF1) & \textbf{Clu} (NMI) & \textbf{Sim@5} \\
        \midrule
        
        \multirow{6}{*}{\texttt{Layers}}
        & \multicolumn{3}{c|}{PAP}  & \multicolumn{3}{c|}{IBI} & \multicolumn{3}{c}{MAM} \\
        & 82.33 {\tiny(6.51)} & 37.01 {\tiny(6.06)} & 85.26 {\tiny(0.56)}
        & 26.50 {\tiny(0.79)} &  0.15 {\tiny(0.04)} & 31.06 {\tiny(0.25)}
        & 51.79 {\tiny(1.24)} & 15.82 {\tiny(4.44)} & 55.72 {\tiny(0.28)} \\

        & \multicolumn{3}{c|}{PSP}  & \multicolumn{3}{c|}{IOI} & \multicolumn{3}{c}{MDM} \\
        & 56.17 {\tiny(5.16)} & 50.42 {\tiny(1.57)} & 66.06 {\tiny(2.88)}
        & 35.63 {\tiny(1.09)} &  1.01 {\tiny(0.53)} & 48.83 {\tiny(1.08)}
        & 35.97 {\tiny(1.28)} &  0.87 {\tiny(0.08)} & 41.28 {\tiny(1.06)} \\

        & \multicolumn{3}{c|}{}  & \multicolumn{3}{c|}{IVI} & \multicolumn{3}{c}{MWM} \\
        & & & 
        & 27.08 {\tiny(0.63)} & 0.14 {\tiny(0.03)} & 29.94 {\tiny(0.33)}
        & 25.26 {\tiny(1.00)} & 0.27 {\tiny(0.09)} & 41.54 {\tiny(1.12)}\\
        \midrule

        \texttt{Features}
        & 73.30 {\tiny(0.00)} & 14.83 {\tiny(4.14)} & 71.75 {\tiny(0.00)}
        & 71.27 {\tiny(0.00)} &  4.11 {\tiny(1.80)} & 84.09 {\tiny(0.00)}
        & 20.18 {\tiny(0.00)} &  0.41 {\tiny(0.03)} & 32.04 {\tiny(0.00)} \\
        
        \bottomrule
        
\end{tabular}
}

\resizebox{\textwidth}{!}{
\begin{tabular}{l|ccc|ccc|ccc}
        \toprule
        & \multicolumn{3}{c|}{\textbf{IMDB}} 
        & \multicolumn{3}{c|}{\textbf{Cora}}
        & \multicolumn{3}{c}{\textbf{CiteSeer}} \\
        
        & \textbf{Clf} (MaF1) & \textbf{Clu} (NMI) & \textbf{Sim@5}
        & \textbf{Clf} (MaF1) & \textbf{Clu} (NMI) & \textbf{Sim@5}
        & \textbf{Clf} (MaF1) & \textbf{Clu} (NMI) & \textbf{Sim@5} \\
        \midrule
        
        \multirow{4}{*}{\texttt{Layers}}
        & \multicolumn{3}{c|}{MAM}  & \multicolumn{3}{c|}{CIT} & \multicolumn{3}{c}{CIT} \\
        & 41.99 {\tiny(2.92)} &  0.48 {\tiny(0.28)} & 45.98 {\tiny(0.30)} 
        & 65.32 {\tiny(3.37)} & 25.70 {\tiny(5.26)} & 75.49 {\tiny(1.37)} 
        & 58.50 {\tiny(3.74)} & 27.36 {\tiny(4.74)} & 63.47 {\tiny(0.82)} \\

        & \multicolumn{3}{c|}{MDM}  & \multicolumn{3}{c|}{KNN} & \multicolumn{3}{c}{KNN} \\
        & 51.50 {\tiny(2.01)} &  5.09 {\tiny(1.36)} & 51.68 {\tiny(0.36)} 
        & 51.87 {\tiny(4.57)} & 16.99 {\tiny(1.01)} & 57.67 {\tiny(1.17)} 
        & 55.93 {\tiny(1.89)} & 24.08 {\tiny(1.49)} & 61.23 {\tiny(0.32)} \\

        \midrule

        \texttt{Features}
        & 56.59 {\tiny(0.00)} &  6.75 {\tiny(2.01)} & 51.64 {\tiny(0.00)} 
        & 55.42 {\tiny(0.00)} & 13.47 {\tiny(0.63)} & 50.10 {\tiny(0.00)} 
        & 58.73 {\tiny(0.00)} & 18.64 {\tiny(2.22)} & 53.48 {\tiny(0.00)} \\
        
        \bottomrule
        
\end{tabular}
}

        
        



        
        

\end{table}

\paragraph{\textbf{Discussion.}} Results are given in Table \ref{tab:no-fusion-results}. In most cases, we observe that different graph layers perform differently in the downstream tasks. For the ACM dataset, the smaller layer PAP performs much better in the node classification and similarity search tasks than the much larger PSP layer. However, the clustering performance is better on the PSP layer. Utilizing the node features also yields a decent performance; however, it is not as good as for the PAP layer. A similar situation happens for the IMDB dataset, where the larger MAM layer performs worse than the MDM layer -- this time in all three downstream tasks. The node features yield the best overall performance. Another case worth noting occurs in the Amazon dataset -- the node features outperform all layer information with about 40 pp difference in the node classification task, 50 pp for the similarity search task, and about 4-40 times better performance on the clustering task. Note that the IBI and IVI layers achieve only 0.15\% and 0.14\% NMI, which are inferior scores. In the case of the artificially constructed Cora and CiteSeer datasets, we observe that the added KNN layers achieve similar performance to the node features, which is expected as these datasets are highly homophilic.

\subsection{Graph-level fusion\label{sec:graph-level-fusion-section}}
\paragraph{\textbf{Methods.}} Information fusion in multiplex graph representation learning can be applied in the graphs themselves. Not much research has been carried out, and the most prominent existing method for graph-level fusion is \texttt{MHGCN} \cite{mhgcn}. The method consists of a GCN backbone trained using a link prediction objective with negative edge sampling. For a $K$-layer multiplex graph, the method jointly learns $K$ edge weights: $\beta_1, \beta_2, \ldots, \beta_K$, which are assigned to each edge in the corresponding graph layer. Finally, the multiplex graph is converted into a homogeneous weighted graph, where the edge weights are summed over the graph layers. For instance, if edge $(u, v)$ exists in three graph layers: $i$, $j$ and $k$, the final edge weight is given as: $\beta^{(u, v)} = \beta_i + \beta_j + \beta_k$. We compare the \texttt{MHGCN} method with an approach where we flatten (\texttt{Flattened}) the input graph into a non-weighted homogeneous graph with multi-edges, i.e., $\mathcal{G} = (\mathcal{V}, \mathcal{E}_1, \mathcal{E}_2, \ldots, \mathcal{E}_K, \mathbf{X}) \to G_\text{flattened} = (\mathcal{V}, \mathcal{E}_\Sigma, \mathbf{X})$, where $\mathcal{E}_\Sigma = \mathcal{E}_1 \cup \mathcal{E}_2 \cup \ldots \cup \mathcal{E}_K$ is the multi-set of edges (an edge between two nodes might occur in several graph layers). We evaluate the performance of several popular node representation learning methods for homogenous graphs -- the supervised \texttt{GCN} and \texttt{GAT} models, the unsupervised DeepWalk (\texttt{DW}) and \texttt{DGI} models. 

\paragraph{\textbf{Discussion.}} Results are given in Table \ref{tab:graph-level-fusion-results}. The \texttt{MHGCN} method did not perform well on the ACM, Amazon and IMDB datasets compared to baseline approaches (no fusion). This might be related to the overall high number of edges (approx. 2M and 1.3M, respectively) combined with the link prediction objective -- mining the negative edges might be troublesome. We leave the evaluation of a modified model training objective as future work. On the contrary, for the Freebase dataset, we observe an approx. 11 pp improvement in the node classification task along with substantial improvement in the similarity search task and comparable results for the clustering task. Similarly, Cora and CiteSeer also benefit from the learnable graph-level fusion approach, which increases performance by up to 15 pp. When it comes to the \texttt{Flattened} graph methods, we firstly observe a consistently better performance for the supervised models compared to unsupervised ones, which is related to the direct access to node labels. However, the \texttt{DW} and \texttt{DGI} methods achieve similar results to \texttt{MHGCN} on ACM and Amazon, while for other datasets, the results are 5-10 pp worse.

\begin{table}[ht]
    \centering

    \caption{Downstream tasks performance of \textbf{graph-level fusion methods}.} 
    \label{tab:graph-level-fusion-results}

\resizebox{\textwidth}{!}{
\begin{tabular}{ll|ccc|ccc|ccc}
        \toprule
        & & \multicolumn{3}{c|}{\textbf{ACM}} 
        & \multicolumn{3}{c|}{\textbf{Amazon}}
        & \multicolumn{3}{c}{\textbf{Freebase}} \\
      
        & & \textbf{Clf} (MaF1) & \textbf{Clu} (NMI) & \textbf{Sim@5}
        & \textbf{Clf} (MaF1) & \textbf{Clu} (NMI) & \textbf{Sim@5}
        & \textbf{Clf} (MaF1) & \textbf{Clu} (NMI) & \textbf{Sim@5} \\
        \midrule

        \rcl 
        & \texttt{GCN}
        & 70.19 {\tiny(0.22)} & 50.12 {\tiny(0.35)} & 73.38 {\tiny(0.23)} 
        & 27.84 {\tiny(0.17)} & 0.81 {\tiny(0.11)} & 27.88 {\tiny(0.20)} 
        & 60.70 {\tiny(0.55)} & 18.71 {\tiny(1.20)} & 56.91 {\tiny(0.43)} \\

        \rcl
        & \texttt{GAT}
        & 70.34 {\tiny(2.89)} & 44.35 {\tiny(2.10)} & 72.17 {\tiny(1.16)} 
        & 25.88 {\tiny(2.06)} & 0.20 {\tiny(0.10)} & 30.08 {\tiny(1.02)} 
        & 57.22 {\tiny(1.24)} & 16.54 {\tiny(1.26)} & 56.80 {\tiny(0.42)} \\

        \cline{2-11}

        \rcl
        & \texttt{DW}
        & 64.25 {\tiny(0.64)} & 37.53 {\tiny(1.82)} & 62.98 {\tiny(0.79)} 
        & 24.45 {\tiny(0.39)} & 0.04 {\tiny(0.01)} & 26.79 {\tiny(0.39)} 
        & 49.20 {\tiny(1.55)} & 17.01 {\tiny(0.47)} & 53.67 {\tiny(0.79)} \\

        \multirow{-4}{*}{\rotatebox{90}{\ccl\texttt{Flattened}}}
        & \texttt{DGI}\ccl
        & 59.54 {\tiny(0.78)}\ccl & 36.07 {\tiny(7.00)}\ccl & 72.85 {\tiny(0.55)} \ccl
        & 25.60 {\tiny(0.92)}\ccl & 0.12 {\tiny(0.02)}\ccl & 28.38 {\tiny(0.15)} \ccl
        & 51.22 {\tiny(0.57)}\ccl & 17.86 {\tiny(0.79)}\ccl & 55.72 {\tiny(0.51)} \ccl\\
        
        \midrule

        \multicolumn{2}{l|}{\texttt{MHGCN}}
        & 63.35 {\tiny(0.88)} & 38.89 {\tiny(0.00)} & 71.05 {\tiny(3.13)} 
        & 23.49 {\tiny(1.69)} & 0.04 {\tiny(0.01)} & 29.44 {\tiny(0.17)} 
        & 62.75 {\tiny(0.20)} & 14.48 {\tiny(1.59)} & 60.84 {\tiny(0.61)} \\
      
        \bottomrule
      
\end{tabular}
}

\resizebox{\textwidth}{!}{
\begin{tabular}{ll|ccc|ccc|ccc}
        \toprule
        & & \multicolumn{3}{c|}{\textbf{IMDB}} 
        & \multicolumn{3}{c|}{\textbf{Cora}}
        & \multicolumn{3}{c}{\textbf{CiteSeer}} \\
      
        & & \textbf{Clf} (MaF1) & \textbf{Clu} (NMI) & \textbf{Sim@5}
        & \textbf{Clf} (MaF1) & \textbf{Clu} (NMI) & \textbf{Sim@5}
        & \textbf{Clf} (MaF1) & \textbf{Clu} (NMI) & \textbf{Sim@5} \\
        \midrule

        \rcl
        & \texttt{GCN}
        & 57.04 {\tiny(0.50)} & 10.69 {\tiny(0.21)} & 49.01 {\tiny(1.03)} 
        & 68.76 {\tiny(0.45)} & 41.11 {\tiny(0.82)} & 66.65 {\tiny(0.33)} 
        & 64.27 {\tiny(0.22)} & 39.93 {\tiny(0.75)} & 62.28 {\tiny(0.73)} \\

        \rcl
        & \texttt{GAT}
        & 54.81 {\tiny(0.67)} & 10.11 {\tiny(0.49)} & 46.90 {\tiny(0.61)} 
        & 65.29 {\tiny(1.68)} & 36.99 {\tiny(1.94)} & 59.94 {\tiny(1.18)} 
        & 60.53 {\tiny(1.84)} & 32.84 {\tiny(3.10)} & 60.17 {\tiny(1.37)} \\

        \cline{2-11}

        \rcl
        & \texttt{DW}
        & 38.09 {\tiny(0.78)} & 0.94 {\tiny(0.19)} & 38.11 {\tiny(0.60)} 
        & 61.50 {\tiny(1.25)} & 39.71 {\tiny(1.00)} & 53.86 {\tiny(1.24)} 
        & 54.16 {\tiny(1.35)} & 40.38 {\tiny(1.07)} & 53.74 {\tiny(0.32)} \\

        \multirow{-4}{*}{\rotatebox{90}{\ccl\texttt{Flattened}}} & \texttt{DGI}\ccl
        & 40.51 {\tiny(4.44)}\ccl &  0.27 {\tiny(0.08)}\ccl & 48.84 {\tiny(0.49)}\ccl 
        & 58.09 {\tiny(3.60)}\ccl & 21.53 {\tiny(3.48)}\ccl & 62.57 {\tiny(0.64)}\ccl 
        & 60.43 {\tiny(2.25)}\ccl & 29.08 {\tiny(2.38)}\ccl & 63.38 {\tiny(0.71)}\ccl \\
        
        \midrule

        \multicolumn{2}{l|}{\texttt{MHGCN}}
        & 49.77 {\tiny(0.83)} & 2.69 {\tiny(0.78)} & 50.07 {\tiny(0.56)} 
        & 71.22 {\tiny(1.17)} & 35.61 {\tiny(1.70)} & 67.09 {\tiny(0.25)} 
        & 62.55 {\tiny(0.84)} & 39.64 {\tiny(2.13)} & 63.02 {\tiny(0.71)} \\
      
        \bottomrule
      
\end{tabular}
}

\end{table}

\subsection{Prediction-level fusion}
\paragraph{\textbf{Methods.}} In this scenario, we assume that precomputed node embedding vectors exist for each multiplex graph layer. We utilize the layer-wise node representations of the \texttt{DGI} method from the baseline experiments in Section \ref{sec:baseline-approches}. Based on those embeddings and the node labels, we train logistic regression classifiers and combine their predictions to obtain a prediction-level information fusion scheme. In particular, we validate two approaches -- \texttt{soft} and \texttt{hard} ensemble voting. In the \texttt{hard} voting case, we first obtain predicted node labels from each classifier and find the most occurring label (also known as majority voting). In the \texttt{soft} case, we first obtain the class probabilities, average them over all classifiers, and select the label with the highest probability. Note that such a scheme is applicable only in the node classification task.  

\paragraph{\textbf{Discussion.}} Results are given in Table \ref{tab:prediction-level-fusion-results}. We observe that the \texttt{soft} approach performs consistently better on all datasets compared to \texttt{hard}. One might lose too much information about the actual class distribution and model confidence by discarding the actual probabilities. Compared to baseline methods and graph-level fusion, we do not observe a significant improvement, yet sometimes, the performance is worse than just utilizing a single graph layer. The information fusion occurs too late in the graph processing pipeline, making discovering complex relationships in the multiplex data harder.

\begin{table}[ht]
    \centering

    \caption{Downstream tasks performance of \textbf{prediction-level fusion methods}.} 
    \label{tab:prediction-level-fusion-results}

\resizebox{\textwidth}{!}{
\begin{tabular}{ll|c|c|c|c|c|c}
        \toprule
        & & \textbf{ACM} 
        & \textbf{Amazon}
        & \textbf{Freebase}
        & \textbf{IMDB} 
        & \textbf{Cora}
        & \textbf{CiteSeer} \\
      
        & & \textbf{Clf} (MaF1)
        & \textbf{Clf} (MaF1) 
        & \textbf{Clf} (MaF1) 
        & \textbf{Clf} (MaF1)
        & \textbf{Clf} (MaF1) 
        & \textbf{Clf} (MaF1)  \\
        \midrule

        \rcl
        & \texttt{soft}
        & 81.95 {\tiny(6.23)} & 36.20 {\tiny(1.01)} & 51.08 {\tiny(1.17)} 
        & 49.73 {\tiny(3.02)} & 65.38 {\tiny(4.12)} & 60.40 {\tiny(2.71)} \\

        \multirow{-2}{*}{\rotatebox{90}{\ccl\texttt{Vote}}} & \texttt{hard}\ccl
        & 58.32 {\tiny(6.81)}\ccl & 31.34 {\tiny(0.89)}\ccl & 45.18 {\tiny(1.04)}\ccl
        & 42.13 {\tiny(3.51)}\ccl & 61.68 {\tiny(2.77)}\ccl & 57.32 {\tiny(1.98)}\ccl \\
        
        \bottomrule
      
\end{tabular}
}

\end{table}

\subsection{GNN-level fusion}
\paragraph{\textbf{Methods.}} This case is where most existing methods apply the fusion mechanism. It seems a natural choice to perform information fusion as a part of the representation learning model. \texttt{DMGI} \cite{dmgi} applies a GCN backbone with the DGI objective at each graph layer and learns the fused node representations as a trainable embedding lookup matrix. The additional loss term minimizes the distance to positive layer-wise embeddings (original graph) and maximizes the distance to negative ones (corrupted graph). \texttt{HDGI} \cite{hdgi} employs a similar approach but uses a semantic attention mechanism to fuse the layer-wise node embeddings. Instead of applying the DGI loss to each graph layer, the model applies it to the fused embeddings. \texttt{S$^2$MGRL} \cite{s2mgrl} builds upon GBT \cite{BIELAK2022109631} and extends it to multiplex networks. Each graph layer is processed by an MLP, followed by a GCN block. The loss function is twofold: (1) a Barlow Twins (BT) loss \cite{zbontar2021barlow} term between the outputs of the MLPs and GCNs, and (2) the BT loss computed between all pairs of GCN outputs. Contrary to previous approaches, S$^2$MGRL is not fully un-/self-supervised. It trains an attention mechanism on pairs of frozen layer-wise embeddings and node labels to fuse node embeddings. We propose extensions to the above methods: \texttt{F(GBT, $\ast$)} and \texttt{F(DGI, $\ast$)}, where $\ast$ denotes the fusion method (attention \texttt{Att} or concatenation with linear projection \texttt{CL}). For \texttt{F(GBT, $\ast$)} we employ a similar scheme as \texttt{S$^2$MGRL}, but instead of computing the BT loss for all graph layer pairs, we compute the BT loss between each graph layer output and the fused node embedding (which scales linearly). For the \texttt{F(DGI, $\cdot$)}, we apply the DGI loss at each graph layer, and additionally on the fused node embeddings (of the original and corrupted graph). Such settings are both inductive and self-supervised.

\begin{table}[ht]
    \centering

    \caption{Downstream tasks performance of \textbf{GNN-level fusion methods}.} 
    \label{tab:gnn-level-fusion-results}

\resizebox{\textwidth}{!}{
\begin{tabular}{l|ccc|ccc|ccc}
        \toprule
        & \multicolumn{3}{c|}{\textbf{ACM}} 
        & \multicolumn{3}{c|}{\textbf{Amazon}}
        & \multicolumn{3}{c}{\textbf{Freebase}} \\
      
        & \textbf{Clf} (MaF1) & \textbf{Clu} (NMI) & \textbf{Sim@5}
        & \textbf{Clf} (MaF1) & \textbf{Clu} (NMI) & \textbf{Sim@5}
        & \textbf{Clf} (MaF1) & \textbf{Clu} (NMI) & \textbf{Sim@5} \\
        \midrule
      
        \texttt{DMGI}
        & 87.76 {\tiny(0.82)} & 57.29 {\tiny(4.47)} & 87.94 {\tiny(0.52)} 
        & 73.15 {\tiny(1.85)} & 36.11 {\tiny(2.60)} & 79.27 {\tiny(0.59)} 
        & 49.79 {\tiny(1.97)} & 15.36 {\tiny(0.49)} & 54.83 {\tiny(2.16)} \\

        \texttt{HDGI}
        & 85.01 {\tiny(2.47)} & 56.33 {\tiny(9.53)} & 87.28 {\tiny(0.64)} 
        & 37.16 {\tiny(3.53)} &  1.21 {\tiny(0.96)} & 44.40 {\tiny(2.25)} 
        & 47.00 {\tiny(2.55)} & 16.58 {\tiny(0.51)} & 55.29 {\tiny(0.53)} \\

        \texttt{S$^2$MGRL}
        & 86.14 {\tiny(2.14)} & 43.02 {\tiny(25.10)} & 85.79 {\tiny(1.04)} 
        & 50.99 {\tiny(3.78)} & 3.91 {\tiny(1.86)} & 58.27 {\tiny(2.80)} 
        & 47.63 {\tiny(7.63)} & 7.22 {\tiny(3.79)} & 57.41 {\tiny(1.46)} \\

        \midrule

        \rcl
        \texttt{F(GBT,Att)}
        & 74.52 {\tiny(3.22)} & 22.90 {\tiny(12.59)} & 78.45 {\tiny(2.04)} 
        & 48.06 {\tiny(2.64)} &  0.76 {\tiny(0.17)} &  52.51 {\tiny(0.85)} 
        & 49.20 {\tiny(2.48)} &  2.01 {\tiny(1.21)} &  55.53 {\tiny(1.57)} \\

        \rcl
        \texttt{F(GBT,CL)}
        & 81.37 {\tiny(2.18)} & 30.54 {\tiny(23.06)} & 83.41 {\tiny(1.51)} 
        & 50.54 {\tiny(4.17)} &  1.41 {\tiny(1.66)} &  52.22 {\tiny(3.07)} 
        & 48.26 {\tiny(2.07)} &  3.02 {\tiny(2.53)} &  55.51 {\tiny(1.12)} \\

        \midrule

        \rcl
        \texttt{F(DGI,Att)}
        & 86.68 {\tiny(2.16)} & 67.57 {\tiny(4.36)} & 89.05 {\tiny(0.45)} 
        & 31.17 {\tiny(3.54)} &  0.39 {\tiny(0.16)} & 44.86 {\tiny(2.97)} 
        & 49.79 {\tiny(0.98)} & 16.09 {\tiny(1.17)} & 57.50 {\tiny(0.59)} \\

        \rcl
        \texttt{F(DGI,CL)}
        & 84.11 {\tiny(4.30)} & 61.13 {\tiny(6.28)} & 89.21 {\tiny(0.29)} 
        & 34.24 {\tiny(2.68)} &  0.28 {\tiny(0.10)} & 42.28 {\tiny(3.38)} 
        & 49.21 {\tiny(0.97)} &  0.80 {\tiny(0.57)} & 52.97 {\tiny(0.95)} \\
        
        \bottomrule
      
\end{tabular}
}

\resizebox{\textwidth}{!}{
\begin{tabular}{l|ccc|ccc|ccc}
        \toprule
        & \multicolumn{3}{c|}{\textbf{IMDB}} 
        & \multicolumn{3}{c|}{\textbf{Cora}}
        & \multicolumn{3}{c}{\textbf{CiteSeer}} \\
      
        & \textbf{Clf} (MaF1) & \textbf{Clu} (NMI) & \textbf{Sim@5}
        & \textbf{Clf} (MaF1) & \textbf{Clu} (NMI) & \textbf{Sim@5}
        & \textbf{Clf} (MaF1) & \textbf{Clu} (NMI) & \textbf{Sim@5} \\
        \midrule
      
        \texttt{DMGI}
        & 60.81 {\tiny(0.66)} & 19.70 {\tiny(0.49)} & 60.05 {\tiny(0.63)} 
        & 73.55 {\tiny(0.56)} & 42.30 {\tiny(2.70)} & 72.78 {\tiny(0.83)} 
        & 66.13 {\tiny(1.75)} & 43.37 {\tiny(0.59)} & 66.84 {\tiny(0.56)} \\

        \texttt{HDGI}
        & 52.60 {\tiny(1.70)} & 13.66 {\tiny(1.42)} & 55.45 {\tiny(1.36)} 
        & 61.45 {\tiny(2.53)} & 34.18 {\tiny(3.00)} & 67.43 {\tiny(2.80)} 
        & 60.54 {\tiny(1.18)} & 40.01 {\tiny(2.18)} & 64.06 {\tiny(1.54)} \\

        \texttt{S$^2$MGRL}
        & 44.23 {\tiny(8.44)} & 1.36 {\tiny(1.76)} & 48.84 {\tiny(2.55)} 
        & 61.03 {\tiny(7.64)} & 19.99 {\tiny(11.78)} & 67.86 {\tiny(1.71)} 
        & 52.66 {\tiny(3.98)} & 20.96 {\tiny(8.81)} & 62.30 {\tiny(0.61)} \\
        
        \midrule

        \rcl
        \texttt{F(GBT,Att)}
        & 53.70 {\tiny(0.95)} &  2.00 {\tiny(2.06)} & 50.39 {\tiny(0.42)} 
        & 68.70 {\tiny(2.22)} & 22.01 {\tiny(3.50)} & 68.60 {\tiny(1.01)} 
        & 59.30 {\tiny(2.40)} & 19.95 {\tiny(6.32)} & 61.36 {\tiny(1.20)} \\

        \rcl
        \texttt{F(GBT,CL)}
        & 56.30 {\tiny(1.08)} &  0.53 {\tiny(0.39)} & 52.59 {\tiny(0.48)} 
        & 70.57 {\tiny(1.61)} & 21.61 {\tiny(1.48)} & 69.87 {\tiny(1.21)} 
        & 60.39 {\tiny(2.24)} & 18.61 {\tiny(5.33)} & 62.54 {\tiny(0.98)} \\

        \midrule

        \rcl
        \texttt{F(DGI,Att)}
        & 55.56 {\tiny(1.53)} & 11.85 {\tiny(3.88)} & 56.59 {\tiny(0.74)} 
        & 66.44 {\tiny(1.01)} & 35.18 {\tiny(0.46)} & 68.18 {\tiny(0.71)} 
        & 62.25 {\tiny(2.04)} & 36.58 {\tiny(3.17)} & 62.58 {\tiny(0.52)} \\

        \rcl
        \texttt{F(DGI,CL)}
        & 52.29 {\tiny(3.40)} &  1.03 {\tiny(1.58)} & 53.83 {\tiny(1.02)} 
        & 62.33 {\tiny(3.68)} & 13.51 {\tiny(4.36)} & 66.40 {\tiny(1.15)} 
        & 60.88 {\tiny(1.49)} & 17.55 {\tiny(1.49)} & 64.08 {\tiny(0.74)} \\
        
        \bottomrule
      
\end{tabular}
}

\end{table}

\paragraph{\textbf{Discussion.}} Results are given in Table \ref{tab:gnn-level-fusion-results}. We observe that the lookup embedding-based fusion approach of the \texttt{DMGI} method consistently delivers the best results on the node classification task compared to the other two approaches. A similar observation can be made for the clustering and similarity search task except for the Freebase dataset, where the performance is alike. Compared to previous fusion scenarios, the results of \texttt{DMGI} are best overall on all datasets except for Freebase, where the model achieves slightly worse results. One could argue that such lookup embedding is the best fusion mechanism, but please note that it does not allow us to easily obtain embeddings for newly added nodes or changes in the graph structure (it is transductive). \texttt{HDGI}'s attention mechanism proves to perform similarly to \texttt{DMGI} on ACM and Freebase, worse on IMDB, Cora and CiteSeer, but on the Amazon dataset, we observe an approx. 35 pp difference in the node classification and similarity search tasks. The utilization of the BT objective in \texttt{S$^2$MGRL} achieves comparable performance to the aforementioned methods on ACM, Freebase and Cora, but on Amazon and IMDB, the difference is more substantial. For our proposed extensions, \texttt{F(GBT, $\ast$)} and \texttt{F(DGI, $\ast$)}, we observe that both variants provide competitive performance to existing multiplex approaches. If we compare the results to \texttt{HDGI}, which is the only model that is both inductive and self-supervised, the performance gains are even greater. For instance, 56\% vs. 52\% MaF1 on IMDB, 70\% vs. 61\% MaF1 on Cora, or even 50\% vs. 37\% MaF1 on Amazon.

\subsection{Embedding-level fusion}\label{sec:embedding-level-fusion-section}

\paragraph{\textbf{Methods.}} A basic fusion approach often presented in multiplex representation learning papers is first to precompute node embeddings for each graph layer separately and then fuse them by computing the average vectors (over all graph layers). We denote that approach as \texttt{Mean} and use it in combination with a supervised \texttt{GCN}, \texttt{GAT} and unsupervised DeepWalk (\texttt{DW}) and \texttt{DGI} models (same set of models as in the flattened graph case -- see Section \ref{sec:graph-level-fusion-section}). We extend these embedding-level (post-hoc) fusion methods by examining other non-trainable fusion operators (i.e., \texttt{Concat}, \texttt{Min}, \texttt{Max}, \texttt{Sum}). We also build models with trainable fusion mechanisms -- i.e., attention (\texttt{Att}), concatenation with a linear projection (\texttt{CL}) and a lookup embedding (\texttt{Lk}). We utilize an MSE or BT loss computed between the layer-wise embedding vectors and the fused output representation.

\begin{table}[ht]
    \centering

    \caption{Downstream tasks performance of \textbf{embedding-level fusion methods}.}
    \label{tab:embedding-level-fusion-results}

\resizebox{\textwidth}{!}{
\begin{tabular}{ll|ccc|ccc|ccc}
        \toprule
        & & \multicolumn{3}{c|}{\textbf{ACM}} 
        & \multicolumn{3}{c|}{\textbf{Amazon}}
        & \multicolumn{3}{c}{\textbf{Freebase}} \\
      
        & & \textbf{Clf} (MaF1) & \textbf{Clu} (NMI) & \textbf{Sim@5}
        & \textbf{Clf} (MaF1) & \textbf{Clu} (NMI) & \textbf{Sim@5}
        & \textbf{Clf} (MaF1) & \textbf{Clu} (NMI) & \textbf{Sim@5} \\
        \midrule
      
        \texttt{GCN} & \multirow{4}{*}{\texttt{Mean}}
        & 84.66 {\tiny(0.16)} & 62.53 {\tiny(0.53)} & 85.61 {\tiny(0.11)} 
        & 45.31 {\tiny(11.90)} & 9.06 {\tiny(8.14)} & 46.05 {\tiny(5.44)} 
        & 58.32 {\tiny(2.69)} & 20.24 {\tiny(1.05)} & 58.07 {\tiny(0.66)} \\
        
        \texttt{GAT} &
        & 85.03 {\tiny(0.98)} & 59.55 {\tiny(4.51)} & 85.36 {\tiny(0.16)} 
        & 35.83 {\tiny(9.66)} &  4.18 {\tiny(6.28)} & 41.13 {\tiny(4.05)} 
        & 50.38 {\tiny(6.87)} & 15.81 {\tiny(3.89)} & 57.44 {\tiny(0.65)} \\

        \cline{1-1}\cline{3-11}
        
        \texttt{DW} &
        & 69.21 {\tiny(0.96)} & 35.63 {\tiny(3.20)} & 64.04 {\tiny(3.23)} 
        & 24.33 {\tiny(0.59)} & 0.04 {\tiny(0.02)} & 26.74 {\tiny(0.53)} 
        & 41.61 {\tiny(2.48)} & 8.75 {\tiny(3.98)} & 43.28 {\tiny(2.37)} \\
        
        \texttt{DGI} &
        & 80.14 {\tiny(8.60)} & 41.12 {\tiny(4.15)} & 85.95 {\tiny(1.04)} 
        & 31.93 {\tiny(0.43)} &  0.36 {\tiny(0.09)} & 42.93 {\tiny(0.58)} 
        & 49.22 {\tiny(2.47)} &  6.57 {\tiny(8.05)} & 54.36 {\tiny(1.42)} \\

        \midrule

        \rcl
        & \texttt{Concat}
        & 82.42 {\tiny(6.51)} & 42.02 {\tiny(4.81)} & 85.90 {\tiny(1.09)} 
        & 36.21 {\tiny(1.02)} &  0.45 {\tiny(0.13)} & 46.42 {\tiny(0.81)} 
        & 50.90 {\tiny(1.22)} &  7.52 {\tiny(7.62)} & 54.82 {\tiny(1.56)} \\

        \rcl
        & \texttt{Min}                                                  
        & 77.80 {\tiny(9.57)} & 42.40 {\tiny(7.76)} & 85.69 {\tiny(0.98)} 
        & 28.22 {\tiny(2.20)} &  0.72 {\tiny(0.04)} & 39.45 {\tiny(0.33)} 
        & 50.60 {\tiny(1.44)} &  8.51 {\tiny(7.69)} & 55.35 {\tiny(0.80)} \\

        \rcl
        & \texttt{Max}                                                  
        & 82.21 {\tiny(6.30)} & 40.35 {\tiny(4.88)} & 85.94 {\tiny(1.06)} 
        & 28.49 {\tiny(1.00)} &  0.72 {\tiny(0.09)} & 37.45 {\tiny(0.51)} 
        & 49.89 {\tiny(2.58)} &  9.08 {\tiny(8.84)} & 54.66 {\tiny(0.82)} \\

        \multirow{-4}{*}{\ccl\texttt{DGI}} & \texttt{Sum}\ccl  
        & 82.44 {\tiny(6.42)}\ccl & 41.12 {\tiny(4.15)}\ccl & 85.95 {\tiny(1.04)}\ccl 
        & 31.96 {\tiny(0.44)}\ccl &  0.36 {\tiny(0.09)}\ccl & 42.93 {\tiny(0.58)}\ccl 
        & 50.53 {\tiny(1.53)}\ccl &  6.57 {\tiny(8.05)}\ccl & 54.36 {\tiny(1.42)}\ccl \\

        \midrule

        \rcl
        & \texttt{Att,BT}
        & 80.37 {\tiny(9.83)} & 50.97 {\tiny(6.63)} & 86.17 {\tiny(0.90)} 
        & 31.60 {\tiny(0.48)} &  0.31 {\tiny(0.06)} & 42.04 {\tiny(1.25)} 
        & 49.67 {\tiny(2.03)} &  8.07 {\tiny(7.38)} & 54.48 {\tiny(1.13)} \\

        \rcl
        & \texttt{Att,MSE}                                                      
        & 80.12 {\tiny(8.57)} & 41.48 {\tiny(4.60)} & 85.95 {\tiny(1.02)} 
        & 31.95 {\tiny(0.43)} &  0.39 {\tiny(0.08)} & 42.92 {\tiny(0.58)} 
        & 49.24 {\tiny(2.50)} &  6.57 {\tiny(8.06)} & 54.37 {\tiny(1.38)} \\

        \rcl
        & \texttt{CL,BT}                                                        
        & 85.39 {\tiny(1.79)} & 36.26 {\tiny(6.87)} & 86.10 {\tiny(0.92)} 
        & 30.25 {\tiny(1.05)} &  0.21 {\tiny(0.15)} & 34.17 {\tiny(2.03)} 
        & 50.24 {\tiny(2.01)} &  7.77 {\tiny(3.56)} & 54.23 {\tiny(0.66)} \\

        \rcl
        & \texttt{CL,MSE}                                                       
        & 80.17 {\tiny(9.18)} & 39.82 {\tiny(3.85)} & 85.85 {\tiny(0.73)} 
        & 32.43 {\tiny(0.93)} &  0.31 {\tiny(0.14)} & 42.30 {\tiny(1.00)} 
        & 49.04 {\tiny(2.55)} &  6.36 {\tiny(8.04)} & 54.00 {\tiny(1.08)} \\

        \rcl
        & \texttt{Lk,BT}                                                        
        & 82.71 {\tiny(1.03)} & 43.93 {\tiny(7.08)} & 83.65 {\tiny(1.83)} 
        & 27.86 {\tiny(0.65)} &  0.13 {\tiny(0.05)} & 33.75 {\tiny(1.35)} 
        & 45.18 {\tiny(1.12)} &  8.44 {\tiny(6.00)} & 53.12 {\tiny(0.87)} \\
        
        \multirow{-6}{*}{\ccl\texttt{DGI}} & \texttt{Lk,MSE}\ccl
        & 33.59 {\tiny(0.64)}\ccl & 0.09 {\tiny(0.01)}\ccl & 32.67 {\tiny(0.28)}\ccl 
        & 24.67 {\tiny(0.51)}\ccl & 0.05 {\tiny(0.01)}\ccl & 26.75 {\tiny(0.18)}\ccl 
        & 33.08 {\tiny(2.63)}\ccl & 0.22 {\tiny(0.08)}\ccl & 36.98 {\tiny(0.84)}\ccl \\
        \bottomrule
      
\end{tabular}
}

\resizebox{\textwidth}{!}{
\begin{tabular}{ll|ccc|ccc|ccc}
        \toprule
        & & \multicolumn{3}{c|}{\textbf{IMDB}} 
        & \multicolumn{3}{c|}{\textbf{Cora}}
        & \multicolumn{3}{c}{\textbf{CiteSeer}} \\
      
        & & \textbf{Clf} (MaF1) & \textbf{Clu} (NMI) & \textbf{Sim@5}
        & \textbf{Clf} (MaF1) & \textbf{Clu} (NMI) & \textbf{Sim@5}
        & \textbf{Clf} (MaF1) & \textbf{Clu} (NMI) & \textbf{Sim@5} \\
        \midrule
      
        \texttt{GCN} & \multirow{4}{*}{\texttt{Mean}}
        & 61.63 {\tiny(0.83)} & 16.39 {\tiny(0.82)} & 55.41 {\tiny(0.47)} 
        & 76.88 {\tiny(1.44)} & 48.40 {\tiny(4.13)} & 74.73 {\tiny(1.50)} 
        & 63.50 {\tiny(0.86)} & 39.51 {\tiny(1.79)} & 63.12 {\tiny(0.77)} \\
        
        \texttt{GAT} & 
        & 54.57 {\tiny(4.41)} &  6.46 {\tiny(2.00)} & 51.14 {\tiny(1.28)} 
        & 73.94 {\tiny(3.18)} & 46.98 {\tiny(5.84)} & 72.52 {\tiny(2.01)} 
        & 64.10 {\tiny(0.72)} & 39.20 {\tiny(2.42)} & 64.52 {\tiny(1.09)} \\

        \cline{1-1}\cline{3-11}
        
        \texttt{DW} & 
        & 34.02 {\tiny(0.80)} & 0.11 {\tiny(0.05)} & 35.03 {\tiny(0.54)} 
        & 54.89 {\tiny(2.54)} & 34.13 {\tiny(1.98)} & 49.82 {\tiny(1.61)} 
        & 44.32 {\tiny(1.41)} & 27.10 {\tiny(2.49)} & 43.33 {\tiny(1.30)} \\

        \texttt{DGI} & 
        & 49.05 {\tiny(3.22)} &  0.80 {\tiny(0.86)} & 51.64 {\tiny(0.59)} 
        & 62.63 {\tiny(4.33)} & 22.96 {\tiny(1.42)} & 69.18 {\tiny(2.55)} 
        & 58.37 {\tiny(2.34)} & 27.61 {\tiny(1.48)} & 64.01 {\tiny(0.87)} \\

        \midrule

        \rcl
        & \texttt{Concat}
        & 49.72 {\tiny(2.93)} &  0.61 {\tiny(0.50)} & 52.15 {\tiny(0.43)} 
        & 65.51 {\tiny(4.09)} & 22.84 {\tiny(0.72)} & 70.41 {\tiny(2.44)} 
        & 60.62 {\tiny(2.49)} & 28.82 {\tiny(2.29)} & 64.47 {\tiny(1.04)} \\

        \rcl
        & \texttt{Min}                                                  
        & 49.20 {\tiny(2.74)} &  0.63 {\tiny(0.51)} & 51.48 {\tiny(0.86)} 
        & 63.81 {\tiny(6.87)} & 23.90 {\tiny(5.20)} & 69.86 {\tiny(1.75)} 
        & 58.96 {\tiny(4.59)} & 30.47 {\tiny(2.45)} & 64.36 {\tiny(0.58)} \\

        \rcl
        & \texttt{Max}                                                  
        & 47.27 {\tiny(3.78)} &  0.47 {\tiny(0.25)} & 51.12 {\tiny(0.75)} 
        & 60.24 {\tiny(4.91)} & 19.37 {\tiny(1.81)} & 68.53 {\tiny(3.38)} 
        & 57.62 {\tiny(1.97)} & 23.70 {\tiny(1.83)} & 63.24 {\tiny(0.94)} \\
                                                                        
        \multirow{-4}{*}{\ccl\texttt{DGI}} & \texttt{Sum}\ccl
        & 49.17 {\tiny(3.09)}\ccl &  0.80 {\tiny(0.86)}\ccl & 51.64 {\tiny(0.59)}\ccl 
        & 63.42 {\tiny(4.18)}\ccl & 22.96 {\tiny(1.42)}\ccl & 69.18 {\tiny(2.55)}\ccl 
        & 59.00 {\tiny(2.24)}\ccl & 27.61 {\tiny(1.48)}\ccl & 64.01 {\tiny(0.87)}\ccl \\

        \midrule

        \rcl
        & \texttt{Att,BT}
        & 50.76 {\tiny(0.81)} &  1.46 {\tiny(1.63)} & 51.66 {\tiny(0.73)} 
        & 66.34 {\tiny(3.20)} & 29.10 {\tiny(3.10)} & 70.99 {\tiny(1.11)} 
        & 60.71 {\tiny(2.89)} & 31.49 {\tiny(4.04)} & 64.36 {\tiny(0.93)} \\

        \rcl
        & \texttt{Att,MSE}                                                      
        & 49.05 {\tiny(3.31)} &  0.81 {\tiny(0.89)} & 51.62 {\tiny(0.62)} 
        & 62.56 {\tiny(4.29)} & 22.49 {\tiny(1.45)} & 69.15 {\tiny(2.56)} 
        & 58.37 {\tiny(2.34)} & 27.73 {\tiny(1.43)} & 64.01 {\tiny(0.89)} \\

        \rcl
        & \texttt{CL,BT}                                                        
        & 52.82 {\tiny(1.67)} &  1.51 {\tiny(0.74)} & 50.88 {\tiny(1.33)} 
        & 68.16 {\tiny(2.22)} & 32.67 {\tiny(3.13)} & 70.40 {\tiny(0.78)} 
        & 61.19 {\tiny(2.11)} & 33.92 {\tiny(2.93)} & 62.88 {\tiny(0.47)} \\

        \rcl
        & \texttt{CL,MSE}                                                       
        & 46.95 {\tiny(2.50)} &  0.57 {\tiny(0.55)} & 50.62 {\tiny(1.51)} 
        & 61.61 {\tiny(4.11)} & 22.27 {\tiny(1.65)} & 68.11 {\tiny(2.22)} 
        & 57.91 {\tiny(3.24)} & 27.57 {\tiny(0.71)} & 63.52 {\tiny(0.82)} \\

        \rcl
        & \texttt{Lk,BT}                                                        
        & 50.50 {\tiny(1.78)} &  2.26 {\tiny(2.28)} & 49.40 {\tiny(1.32)} 
        & 68.43 {\tiny(2.03)} & 28.44 {\tiny(3.30)} & 66.39 {\tiny(0.47)} 
        & 54.76 {\tiny(1.54)} & 31.86 {\tiny(0.37)} & 60.92 {\tiny(0.94)} \\
                                                                                
        \multirow{-6}{*}{\ccl\texttt{DGI}} & \texttt{Lk,MSE}\ccl
        & 33.11 {\tiny(0.86)}\ccl &  0.06 {\tiny(0.01)}\ccl & 34.03 {\tiny(0.53)}\ccl 
        & 14.12 {\tiny(1.16)}\ccl &  1.09 {\tiny(0.08)}\ccl & 18.22 {\tiny(0.34)}\ccl 
        & 16.85 {\tiny(1.17)}\ccl &  0.78 {\tiny(0.12)}\ccl & 18.22 {\tiny(0.65)}\ccl \\
        \bottomrule
      
\end{tabular}
}

\end{table}

\paragraph{\textbf{Discussion.}} Results are given in Table \ref{tab:embedding-level-fusion-results}. The supervised \texttt{GCN} and \texttt{GAT} models perform better than the unsupervised ones. However, in some cases, the performance gap is small (e.g., \texttt{GAT} and \texttt{DGI} on Freebase in the node classification task). Regarding other non-trainable operators in conjunction with the \texttt{DGI} model, we observe that \texttt{Concat} often delivers the best performance, however, at the cost of higher-dimensional node embedding vectors. This approach would be infeasible when dealing with multiplex networks with many layers (edge types). The other operators (\texttt{Min}, \texttt{Max}, \texttt{Sum}) perform quite similarly to each other on all datasets. However, on the ACM dataset, we observe that using either \texttt{Max} or \texttt{Sum} aggregation leads to a performance increase of about 2 pp compared to \texttt{Mean} along at a reduced std. Trainable operators also allow for an increase in the performance in downstream tasks in most cases -- on ACM, the best choice is the \texttt{CL, BT} model with approx. 85\% MaF1 and 86\% Sim@5, while for clustering the best score is achieved by \texttt{Att, BT}. On Amazon and Freebase, the results are not significantly different. On IMDB and CiteSeer, we obtain about 2-3pp increase in node classification, and on Cora \texttt{CL, BT} achieves about 68\% MaF1, while outperforming the results of \texttt{DGI, Mean} on clustering and similarity search tasks.

\section{Conclusions}
In this paper, we tackled the problem of node representation learning in multiplex graphs. We proposed a novel taxonomy for categorizing embedding methods based on the location where the information fusion from different graph layers (edge types) occurs. In particular, we identified graph-level, GNN-level, posthoc embedding-level and prediction-level fusion approaches based either on non-trainable basic operators (such as averaging) or more complex trainable ones (such as attention). Most representation learning methods for multiplex graphs are based on the DGI method. We summarize a detailed comparison of methods and fusion approaches in Table \ref{tab:method-comparison}. Combined with our experimental evaluation on three downstream tasks and six multiplex graph datasets, we find the following limitations of existing approaches and propose guidelines for future research (research gaps):
\begin{itemize}
    \item Although various approaches for un-/self-supervised multiplex graph representation learning exist, most are based on the DGI architecture. Self-supervised learning has proved to achieve SoTA performance on homogeneous graph datasets. We propose to evaluate other base architectures.
    \item The attempt to utilize the G-BT architecture on multiplex graphs revealed promising results. However, the final proposed model does not work in a fully unsupervised fashion and requires node labels to train the fusion mechanism.
    \item The \texttt{DMGI} model achieves the best results on most datasets and downstream tasks. However, the utilized lookup embedding-based fusion mechanism does not provide inductive capabilities. This is especially important in real-world applications where the graph data is subject to change.
    \item Inductivity in terms of multiplex graphs can be viewed twofold: (A) a method might be inductive when modifying a single graph layer (addition of nodes or edges) and most GNN-based methods are capable of inferring new embeddings in such scenarios. Exceptions are DeepWalk-based approaches and methods utilizing lookup embeddings as the fusion mechanism. (B) We additionally consider inductivity in terms of adding new multiplex graph layers. Virtually all methods are incapable of such behavior, so adding a new graph layer requires a whole model retraining. Methods based on graph flattening are capable of layer-inductivity, but the resulting node embeddings perform rather poorly in downstream tasks.
    \item  We did not find any attempts to build dedicated GNN layer architectures designed for multiplex graph data. Proposed models instead utilize existing ones and apply them to each graph layer separately, relying on information fusion to be executed in a later step.
    \item Moreover, applying GNNs at each layer separately yields a high number of model parameters. We propose to explore the application of GNN sharing. For instance, this could be achieved by designing new positional encodings for multiplex graph layers.
\end{itemize}

\vspace{-2em}
\begin{table}[ht]
    \centering

    \caption{Comparison of multiplex graph representation learning and fusion methods.}
    \label{tab:method-comparison}

\resizebox{\textwidth}{!}{
\begin{tabular}{ll||c||c|c||c|c|c|c||c}
        \toprule

        \multicolumn{2}{l||}{\textbf{Name}}
        & \textbf{Unsuperivsed?}
        & \multicolumn{2}{c||}{\textbf{Inductive?}} 
        & \multicolumn{4}{c||}{\textbf{Fusion location/level}}
        & \textbf{Fusion type} \\

        & &
        & {\tiny\textbf{Node/edge}}
        & {\tiny\textbf{Layer}} 
        & {\tiny\textbf{Graph}}
        & {\tiny\textbf{GNN}}
        & {\tiny\textbf{Emb.}}
        & {\tiny\textbf{Pred.}}
        & \\
      
        \midrule
        \rcl
        & \texttt{GCN}
        & \no
        & \yes
        & \yes
        & \yes
        & \no
        & \no
        & \no
        & \\

        \rcl
        & \texttt{GAT}
        & \no
        & \yes
        & \yes
        & \yes
        & \no
        & \no
        & \no
        & \\

        \rcl
        & \texttt{DW}
        & \yes
        & \no
        & \no
        & \yes
        & \no
        & \no
        & \no
        & \\

        \rcl
        \multirow{-4}{*}{\rotatebox{90}{\texttt{Flattened}}} & \texttt{DGI}
        & \yes
        & \yes
        & \yes
        & \yes
        & \no
        & \no 
        & \no 
        & \multirow{-4}{*}{graph flattening} \\

        \multicolumn{2}{l||}{\texttt{MHGCN}}
        & \yes
        & \yes
        & \no
        & \yes
        & \no
        & \no 
        & \no 
        & weighted sum of adj. matrices\\

        \midrule

        \multicolumn{2}{l||}{\texttt{DMGI}}
        & \yes
        & \no
        & \no
        & \no
        & \yes
        & \no 
        & \no 
        & lookup\\

        \multicolumn{2}{l||}{\texttt{HDGI}}
        & \yes
        & \yes
        & \no
        & \no
        & \yes
        & \no 
        & \no 
        & semantic attention\\

        \multicolumn{2}{l||}{\texttt{S$^2$MGRL}}
        & \yes (fusion: \no)
        & \yes
        & \no
        & \no
        & \yes
        & \no 
        & \no 
        & supervised attention\\

        \rcl
        \multicolumn{2}{l||}{\texttt{F(GBT, $\ast$)}}
        & \yes
        & \yes
        & \no
        & \no
        & \yes
        & \no 
        & \no 
        &\\

        \rcl
        \multicolumn{2}{l||}{\texttt{F(DGI, $\ast$)}}
        & \yes
        & \yes
        & \no
        & \no
        & \yes
        & \no 
        & \no 
        & \multirow{-2}{*}{attention / concat with linear proj.}\\

        \midrule

        \texttt{GCN} & \multirow{4}{*}{\texttt{Mean}}
        & \no
        & \yes
        & \no
        & \no
        & \no
        & \yes
        & \no 
        & \multirow{4}{*}{mean}\\
        
        \texttt{GAT} &
        & \no
        & \yes
        & \no
        & \no
        & \no
        & \yes
        & \no 
        & \\

        \texttt{DW} &
        & \yes
        & \no
        & \no
        & \no
        & \no
        & \yes
        & \no
        & \\
        
        \texttt{DGI} &
        & \yes
        & \yes
        & \no
        & \no
        & \no
        & \yes
        & \no 
        & \\

        \hline

        \rcl
         & \texttt{Concat}
        & \yes
        & \yes
        & \no
        & \no
        & \no
        & \yes
        & \no 
        & concat \\

        \rcl
        & \texttt{Min}                                                  
        & \yes 
        & \yes
        & \no
        & \no
        & \no
        & \yes
        & \no 
        & min \\

        \rcl
        & \texttt{Max}                                                  
        & \yes
        & \yes
        & \no
        & \no
        & \no
        & \yes
        & \no 
        & max\\

        \rcl
        \multirow{-4}{*}{\texttt{DGI}} & \texttt{Sum}
        & \yes
        & \yes
        & \no
        & \no
        & \no
        & \yes
        & \no 
        & sum \\

        \hline

        \rcl
        & \texttt{Att,BT}
        & \yes
        & \yes
        & \no
        & \no
        & \no
        & \yes
        & \no 
        & \\
                                                                                
        \rcl
        & \texttt{Att,MSE}                                                      
        & \yes
        & \yes
        & \no
        & \no
        & \no
        & \yes
        & \no 
        & \multirow{-2}{*}{attention}\\

        \cline{2-10}
        
        \rcl
        & \texttt{CL,BT}                                                        
        & \yes
        & \yes
        & \no
        & \no
        & \no
        & \yes
        & \no 
        & \\
                                                                                
        \rcl
        & \texttt{CL,MSE}                                                       
        & \yes
        & \yes
        & \no
        & \no
        & \no
        & \yes
        & \no 
        & \multirow{-2}{*}{concat with linear proj.}\\

        \cline{2-10}
        
        \rcl
        & \texttt{Lk,BT}                                                        
        & \yes
        & \no
        & \no
        & \no
        & \no
        & \yes
        & \no 
        &\\
        
        \rcl
        \multirow{-6}{*}{\texttt{DGI}} & \texttt{Lk,MSE}
        & \yes
        & \no
        & \no
        & \no
        & \no
        & \yes
        & \no 
        & \multirow{-2}{*}{lookup} \\

        \midrule

        \rcl
        & \texttt{soft}
        & \no
        & \yes
        & \no
        & \no
        & \no
        & \no 
        & \yes
        & \\

        \rcl
        \multirow{-2}{*}{\texttt{Vote}} & \texttt{hard}
        & \no
        & \yes
        & \no
        & \no
        & \no
        & \no 
        & \yes
        & \multirow{-2}{*}{ensemble voting} \\
        
        \bottomrule
      
\end{tabular}
}

\end{table}

\vspace{-3.5em}
\section*{Acknowledgments}
\vspace{-1em}
The project was partially supported by the National Science Centre, Poland (grant number 2021/41/N/ST6/03694), the European Union under the Horizon Europe grant OMINO (grant number 101086321) and Polish Ministry of Science, as well as statutory funds from the Department of Artificial Intelligence.

\bibliographystyle{splncs04}
\bibliography{main}

\end{document}